\documentclass[conference]{IEEEtran}
\IEEEoverridecommandlockouts
\usepackage{cite}
\usepackage{amsmath,amssymb,amsfonts}
\usepackage{algorithmic}
\usepackage{graphicx}
\usepackage{textcomp}
\usepackage{xcolor}
\usepackage{hyperref} 
\usepackage[utf8]{inputenc}
\usepackage{amssymb}
\def\BibTeX{{\rm B\kern-.05em{\sc i\kern-.025em b}\kern-.08em
    T\kern-.1667em\lower.7ex\hbox{E}\kern-.125emX}}
\begin{document}

\title{AutoOdom: Learning Auto-regressive Proprioceptive Odometry for Legged Locomotion\\
}

\author{
Changsheng Luo\textsuperscript{1}, Yushi Wang\textsuperscript{1}, Wenhan Cai\textsuperscript{2}, Mingguo Zhao\textsuperscript{1}
\thanks{\textsuperscript{1}The authors are with Department of Automation, Tsinghua University, Beijing, China \href{mailto:mgzhao@mail.tsinghua.edu.cn}{mgzhao@mail.tsinghua.edu.cn}}
\thanks{\textsuperscript{2}Wenhan Cai is with Booster Robotics}
}

\maketitle

\begin{abstract}
    Accurate proprioceptive odometry is fundamental for legged robot navigation in GPS-denied and visually degraded environments where conventional visual odometry systems fail. Current approaches face critical limitations: analytical filtering methods suffer from modeling uncertainties and cumulative drift, hybrid learning-filtering approaches remain constrained by their analytical components, while pure learning-based methods struggle with simulation-to-reality transfer and demand extensive real-world data collection. This paper introduces AutoOdom, a novel autoregressive proprioceptive odometry system that overcomes these challenges through an innovative two-stage training paradigm. Stage 1 employs large-scale simulation data to learn complex nonlinear dynamics and rapidly changing contact states inherent in legged locomotion, while Stage 2 introduces an autoregressive enhancement mechanism using limited real-world data to effectively bridge the sim-to-real gap. The key innovation lies in our autoregressive training approach, where the model learns from its own predictions to develop resilience against sensor noise and improve robustness in highly dynamic environments. Comprehensive experimental validation on the Booster T1 humanoid robot demonstrates that AutoOdom significantly outperforms state-of-the-art methods across all evaluation metrics, achieving 57.2\% improvement in absolute trajectory error, 59.2\% improvement in Umeyama-aligned error, and 36.2\% improvement in relative pose error compared to the Legolas baseline. Extensive ablation studies provide critical insights into sensor modality selection and temporal modeling, revealing counterintuitive findings about IMU acceleration data and validating our systematic design choices for robust proprioceptive odometry in challenging locomotion scenarios.
\end{abstract}

\section{Introduction}
Accurate state estimation is crucial for navigation and control of legged robots~\cite{2018arXiv180404811F}, especially in dynamic environments requiring rapid movement. Current precise odometry systems used in robots typically rely on visual sensing~\cite{8421746}~\cite{10421913}~\cite{6096039}~\cite{6153423}, which often fails during intense motions such as jumping, sharp turns, or traversing uneven terrain. In such scenarios, motion blur and lighting variations degrade visual tracking capabilities~\cite{2024arXiv240909287Y}~\cite{paul2024mpvomotionpriorbasedvisual}. This highlights the need for robust proprioceptive odometry systems that can reliably operate using only onboard inertial and joint sensors. However, developing such systems poses significant challenges: rapidly changing contact states during motion result in highly nonlinear dynamics, sensor measurements are prone to substantial noise, and complex foot-ground interactions are difficult to model accurately~\cite{20.500.11850/129873}. These factors collectively demand the design of a noise-resistant odometry estimator that converges quickly while maintaining high accuracy across diverse motion scenarios.

Existing approaches to proprioceptive odometry for legged robots can be broadly categorized into three types: analytical filtering methods, hybrid learning-filtering methods, and pure learning methods. Analytical methods such as the Extended Kalman Filter (EKF) rely on explicit kinematic and contact models~\cite{DBLP:journals/firai/CamurriRNF20}~\cite{10342061}~\cite{9341521}, suffering from modeling uncertainties and numerous assumptions~\cite{9807408}~\cite{20.500.11850/129873}, requiring extensive manual parameter tuning for different robotic platforms and environments. Hybrid approaches attempt to combine learning components with traditional filters, aiming to enhance filtering performance by using neural networks to predict displacement corrections or contact states~\cite{2021arXiv211100789B}~\cite{9134860}. However, these methods still mainly depend on analytical components, thus remaining constrained by the inherent defects of the analytical modules~\cite{10802731}. Although pure learning methods show great potential in handling complex nonlinear dynamics, they demand massive data for training~\cite{17299a5987c541c3861795a72e1679bc}~\cite{2019arXiv190512853Y}~\cite{2018arXiv180202209C}. In practice, real-world data collection faces challenges of time-consuming efforts and equipment limitations in diverse scenarios~\cite{10802731}, while using simulation environments to acquire data introduces significant "sim2real" gaps, which undermine the accuracy during practical deployment~\cite{9308468}.

In this work, we introduce AutoOdom, a novel auto-regressive proprioceptive odometry system that makes two major contributions. First, we develop a two-stage training scheme that integrates simulation and real-world data, where the first stage addresses the challenges of highly nonlinear dynamics and rapidly changing contact states, while the second stage significantly reduces the "Sim2Real" gap. Second, we introduce an auto-regressive training mechanism that reduces the model's sensitivity to noise and enhances the system's robustness in highly dynamic environments.

\begin{figure*}[htbp]
    \centering
    \includegraphics[width=\textwidth]{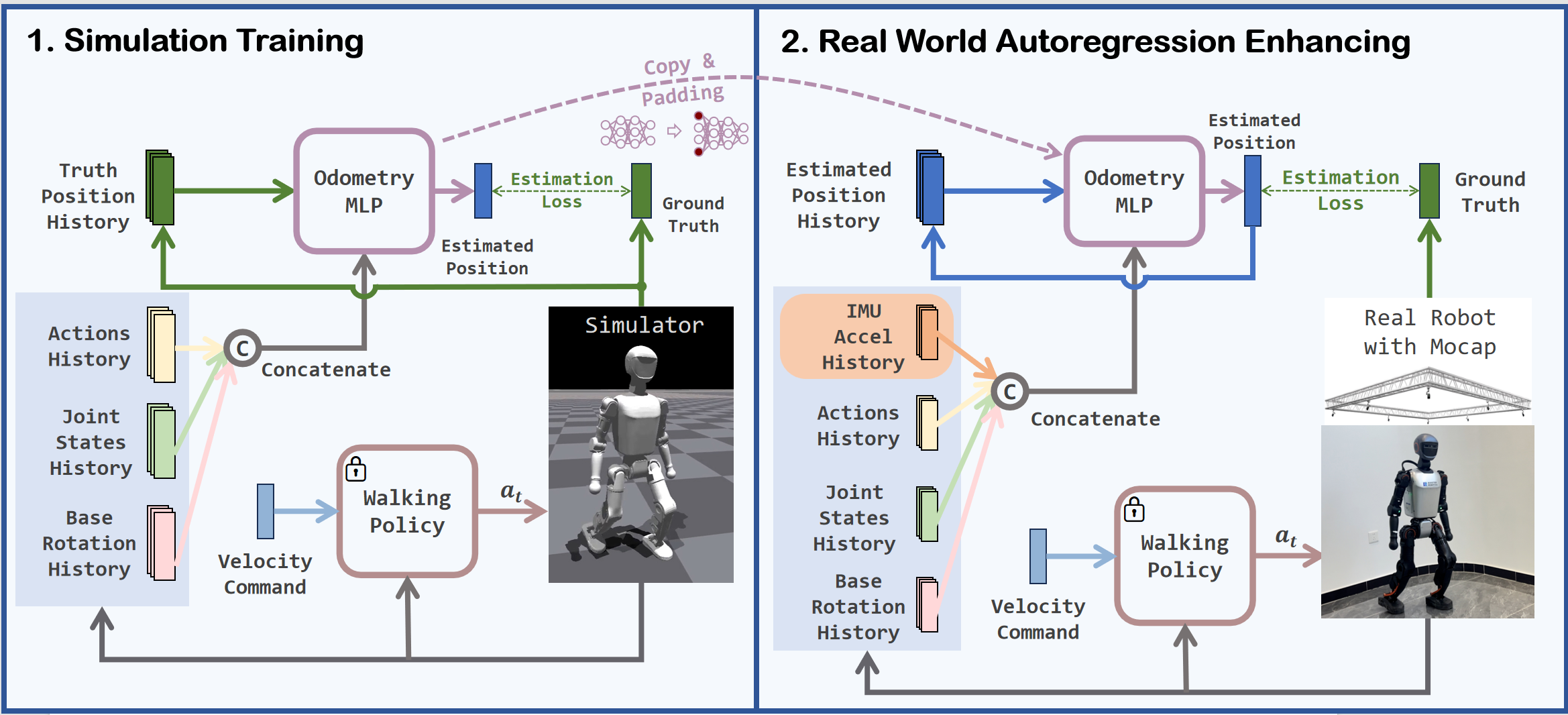}
    \caption{Overview of the AutoOdom system.}
    \label{fig:overview}
\end{figure*}

This method enables precise prediction of pose changes and corresponding variances between consecutive frames for legged robots. Comprehensive experimental results validate that "AutoOdom" outperforms traditional filtering approaches and state-of-the-art simulation-only data-driven legged odometry systems in accuracy within highly dynamic environments.

\section{Related Work}

Proprioceptive odometry for legged robots has evolved significantly over the past decade, driven by the need for robust state estimation in GPS-denied and visually degraded environments. Existing approaches can be broadly categorized into three main paradigms: filtering-based methods, hybrid learning-filtering approaches, and pure learning-based techniques.

\subsection{Filtering-Based Proprioceptive Odometry}

Filtering-based approaches represent the traditional foundation of proprioceptive odometry, leveraging recursive state estimation frameworks such as the Extended Kalman Filter (EKF) or Unscented Kalman Filter (UKF) to fuse measurements from inertial measurement units (IMUs) and joint encoders~\cite{6697236}~\cite{DBLP:journals/firai/CamurriRNF20}~\cite{welch1995kalman}. The classical framework integrates IMU data with leg kinematics under the zero-velocity assumption at ground contact points, predicting state evolution during the prediction step and incorporating leg odometry measurements during the update step. Advanced techniques have addressed the limitations of this assumption, including probabilistic contact estimation~\cite{7815333} and Multi-IMU Proprioceptive Odometry (MIPO)~\cite{10342061}, which achieved order-of-magnitude reductions in position drift through EKF modeling of pivoting contact dynamics.

Despite these advances, filtering-based approaches remain fundamentally constrained by their reliance on analytical models and linearization assumptions. Accumulated drift due to sensor noise and model uncertainties typically exceeds 10\% of traveled distance~\cite{6697236,10342061}, while the complex nonlinear dynamics and contact behaviors of legged robots prove difficult for linearized filters to capture accurately~\cite{17299a5987c541c3861795a72e1679bc}. Recent efforts to mitigate these limitations include multi-sensor fusion strategies, adaptive filtering with dynamic covariance adjustment, and the integration of learning components to predict filter parameters or correct systematic drift~\cite{2021arXiv211100789B}.

\subsection{Hybrid Learning-Filtering Approaches}

Hybrid learning-filtering methods have emerged as a promising paradigm that combines the theoretical rigor of probabilistic filtering with the representational power of deep learning to address the limitations of purely analytical approaches~\cite{DBLP:journals/firai/CamurriRNF20}. The fundamental motivation stems from the recognition that purely model-based methods struggle with unmodeled uncertainties, including contact instability, leg flexibility, and terrain variations, while pure learning-based methods may lack robustness in out-of-distribution scenarios~\cite{17299a5987c541c3861795a72e1679bc}. Hybrid systems bridge this gap by employing learning components to capture residual errors or complex patterns that analytical models fail to represent, while filtering modules provide principled state propagation and fusion of learned corrections.

Key architectural components include neural networks trained to predict contact-related biases, displacement uncertainties, or systematic errors from sequential proprioceptive measurements, with learned displacement estimators using LSTM architectures demonstrating odometry drift reductions of up to 64.93\% compared to baseline filtering methods~\cite{2021arXiv211100789B}. The filtering backbone, typically implemented as an EKF, integrates learned predictions either as pseudo-measurements or through adaptive covariance adjustment, exemplified by TLIO~\cite{9134860}, which employs neural networks to regress 3D displacement and uncertainty estimates for joint estimation of pose, velocity, and IMU biases. By combining data-driven insights into unmodeled dynamics with rigorous Bayesian state estimation, hybrid approaches have demonstrated superior performance across both in-domain and out-of-domain scenarios~\cite{2018arXiv180202209C}, making them particularly valuable for autonomous legged locomotion in unstructured environments.

\subsection{Pure Learning-Based Methods}

Pure learning-based approaches represent a paradigm shift from traditional analytical methods, directly learning complex motion patterns from proprioceptive sensor data without relying on explicit kinematic or dynamic models~\cite{2018arXiv180202209C}. The fundamental innovation lies in eliminating continuous error propagation inherent in classical double integration approaches through window-based formulations. Chen et al.~\cite{2018arXiv180202209C} introduced IONet, which segments inertial measurements into independent temporal windows and employs bidirectional LSTM networks to estimate displacement and heading changes in polar coordinates, demonstrating superior performance across diverse motion patterns. Similar advancements include neural network architectures that fuse kinematic computations with filtered IMU data and actuator measurements, achieving sub-9 cm average localization errors on humanoid platforms~\cite{2019arXiv190512853Y}.

Despite their promising capabilities in handling complex nonlinear dynamics, pure learning-based methods face significant challenges in practical deployment. The requirement for massive training data poses substantial constraints, as real-world data collection is time-consuming and equipment-intensive across diverse scenarios~\cite{10802731}. While simulation environments can provide abundant data, the resulting "sim-to-real" gap significantly undermines deployment accuracy~\cite{9308468}. Additionally, these methods struggle with generalization to unseen terrains and handling sensor bias, requiring careful design of rotation representations and physical constraint integration to enhance robustness~\cite{2018arXiv180202209C}. Despite these limitations, experimental validations across platforms demonstrate their versatility, with reported improvements of 37\% in relative pose error for challenging terrains and 81\% enhancement in localization accuracy over traditional dead reckoning approaches~\cite{2021arXiv211100789B}.

\section{Method}

AutoOdom is designed as a two-stage learning framework that addresses the challenges of proprioceptive odometry through simulation training followed by real-world autoregressive enhancement. The system integrates multiple proprioceptive sensing modalities to capture the complex dynamics of legged locomotion.Our odometry work is built upon Booster Gym, an open-source development platform for the Booster T1 robot~\cite{wang2025boostergymendtoendreinforcement}.

\subsection{Problem Formulation}

At each time step $t$, the system processes a multi-modal observation sequence to predict incremental robot motion. We select a subset of physical quantities from the walking policy's input as proprioceptive measurements for odometry estimation. The input observation $\mathbf{O}_t$ consists of the following components:

\begin{equation}
\mathbf{O}_t = \{\mathbf{A}_t, \mathbf{v}_t^{\text{cmd}}, \boldsymbol{\omega}_t, \mathbf{a}_t, \mathbf{q}_t, \dot{\mathbf{q}}_t, \mathbf{R}_t, \Delta \mathbf{p}_t\}
\end{equation}

where the components are defined as follows:

$\mathbf{A}_t \in \mathbb{R}^{11}$ represents the current actions of the walking policy, including joint commands for 11 actuated joints;

$\mathbf{v}_t^{\text{cmd}} \in \mathbb{R}^{3}$ denotes the current command velocity vector $[v_x^{\text{cmd}}, v_y^{\text{cmd}}, \omega_z^{\text{cmd}}]^{\top}$ representing desired linear and angular motion;

$\boldsymbol{\omega}_t \in \mathbb{R}^{3}$ contains current angular velocity measurements from the gyroscope;

$\mathbf{a}_t \in \mathbb{R}^{3}$ includes current linear acceleration measurements, which are only utilized in the second training stage due to significant discrepancies between simulated and real-world acceleration data;

$\mathbf{q}_t, \dot{\mathbf{q}}_t \in \mathbb{R}^{12}$ capture current joint states including positions and velocities for 12 actuated joints;

$\mathbf{R}_t \in \mathbb{R}^{3 \times 3}$ represents the current rotation matrix from the robot's local frame to the world frame;

 $\Delta \mathbf{p}_t \in \mathbb{R}^{2}$ denotes the differential position change from the current position $\mathbf{p}_t$ to the position 1 second ago $\mathbf{p}_{t-50}$, which corresponds to ground truth in simulation and model estimation in real-world deployment.

The model predicts the incremental position change as follows:

\begin{equation}
\boldsymbol{\Delta}\mathbf{p}_{t+1} = f_{\boldsymbol{\theta}}(\mathbf{O}_{t-H+1:t}) \in \mathbb{R}^{2}
\end{equation}

where $f_{\boldsymbol{\theta}}$ represents the neural network with parameters $\boldsymbol{\theta}$, and $\boldsymbol{\Delta}\mathbf{p}_{t+1} = [\Delta x, \Delta y]^{\top}$ denotes the translational motion increment in the local coordinate frame for the subsequent time interval $\Delta t = 0.02$s. The historical window size $H = 50$ corresponds to 1 second of observation history at 50Hz sampling rate.

\subsection{Stage 1: Simulation-Based Pre-training}

The first stage leverages simulation environments to collect diverse training data spanning various terrains and motion patterns. Using the Isaac Gym simulator, we collect trajectories with robust locomotion policies across environments containing obstacles, flat surfaces, and uneven terrain.

During simulation training, all components of $\mathbf{O}_t$ are obtained as ground truth values from the simulator without any added noise. Specifically, we do not utilize $\mathbf{a}_t^{\text{sim}}$ in the first stage because the acceleration data in simulation is computed through numerical differentiation, which exhibits significant discrepancies compared to actual IMU sensor measurements in the real world.

The simulation environment systematically generates data across varying gait patterns, terrain complexities, and dynamic conditions by executing parallel robot instances. This comprehensive sensor simulation allows the model to initially learn fundamental relationships between proprioceptive inputs and resulting motion without the noise, calibration errors, and hardware limitations present in real-world deployment.

\subsection{Stage 2: Real World Autoregressive Enhancement}

The core innovation lies in the second stage, where we employ an autoregressive enhancement mechanism using limited real-world data. The transition from Stage 1 to Stage 2 involves copy and padding operations to maintain temporal consistency and bridge the simulation-to-reality gap. Rather than relying solely on simulation, this phase fine-tunes the pre-trained model using actual robot trajectories collected from real robots with motion capture systems.

In contrast to the simulation stage, we employ zero-padding techniques to transfer the pre-trained model parameters from Stage 1 to an expanded network architecture in the real-world training phase. The key enhancement involves augmenting the input dimensionality to incorporate real-world IMU acceleration data $\mathbf{a}_t$ as an additional input component, motivated by previous work such as IONet~\cite{2018arXiv180202209C}, which demonstrates that IMU acceleration measurements provide richer dynamic information in real-world scenarios. Furthermore, we implement an autoregressive training mechanism where instead of using ground truth values, we replace $\Delta \mathbf{p}_t$ in $\mathbf{O}_t$ with the model's previous predictions $\Delta \mathbf{p}_t^{\text{pred}}$ computed through cumulative integration of prior estimates. This autoregressive mechanism enables the model to learn from its own predictions at each time step, creating a feedback loop that enhances the current input with historical prediction information and develops resilience to long-term noise relationships, reducing sensitivity to sensor noise and improving robustness in highly dynamic environments. In this second stage, the model processes observation histories $\mathbf{O}_{t-H}, \mathbf{O}_{t-H+1}, \ldots, \mathbf{O}_{t-1}, \mathbf{O}_t$ and predicts incremental motion for the subsequent time interval $\Delta t = 0.02$s based on this comprehensive multi-modal sensor fusion approach. 

The autoregressive mechanism enables the model to learn from its own predictions by feeding back estimated position increments as inputs for subsequent time steps. This creates a closed-loop learning process that reduces sensitivity to simulation-reality gaps while maintaining the robustness gained from large-scale simulation training. The key insight is that the model uses its own estimated position history rather than ground truth, forcing it to learn robust predictions that account for accumulated errors.

\subsection{Training Objective}

Our training objective utilizes a relative pose prediction error that compares predicted and ground truth incremental motions. The key insight is that this error formulation prevents error propagation to future predictions, thereby enabling the creation of independent data points within collected datasets.

The loss function is defined as the mean squared error (MSE) between the predicted and ground truth incremental positions collected by the motion capture system:
\begin{equation}
\mathcal{L}_{\text{MSE}} = \frac{1}{N} \sum_{j=1}^{N} \|\boldsymbol{\Delta}\mathbf{p}_j - \boldsymbol{\Delta}\hat{\mathbf{p}}_j\|_2^2
\end{equation}

where $\boldsymbol{\Delta}\mathbf{p}_j = \mathbf{p}_j - \mathbf{p}_{j-1}$ represents the ground truth position increment and $\boldsymbol{\Delta}\hat{\mathbf{p}}_j = f_{\boldsymbol{\theta}}(\mathbf{O}_{j-H:j})$ represents the predicted increment. This straightforward loss formulation enables effective learning of motion patterns while maintaining computational efficiency during training.

\subsection{Network Architecture}

The network architecture employs fully connected neural networks optimized for real-time deployment. The neural network $f_{\boldsymbol{\theta}}: \mathbb{R}^{d_{\text{in}} \times H} \rightarrow \mathbb{R}^{2}$ maps the multi-modal observation sequence to incremental position predictions, where $d_{\text{in}} \in \{32, 34, \ldots, 46\}$ represents the input feature dimension that varies based on sensor configuration.

We implement multiple architecture variants for comprehensive comparison, including MLP-based and LSTM-based designs:
\begin{align}
f_{\boldsymbol{\theta}}^{\text{MLP}}(\mathbf{O}) &= \text{MLP}(\text{flatten}(\mathbf{O})) \\
f_{\boldsymbol{\theta}}^{\text{LSTM}}(\mathbf{O}) &= \text{Linear}(\text{LSTM}(\mathbf{O})_{-1})
\end{align}

Our subsequent experiments demonstrate that MLP and LSTM architectures achieve comparable performance with negligible differences in accuracy metrics, therefore we ultimately selected MLP as our final network architecture for its superior computational efficiency and lower inference latency in real-time applications. The models typically contain $0.5 \times 10^6$ to $2.0 \times 10^6$ parameters depending on the configuration. 

\section{Experiments}

\subsection{Experimental Setup}

We evaluate AutoOdom on real-world robot trajectories collected across indoor soccer field environments, with deployment on humanoid robotic platforms. The primary experimental validation includes deployment on the Booster T1 humanoid robot from Booster Robotics, demonstrating the versatility of our approach across different locomotion paradigms. The dataset consists of trajectories with an average duration of 20 seconds, covering various walking directions and speeds. The robot operates at 50Hz, providing real-time odometry estimation during dynamic locomotion. Ground truth data is collected using a motion capture system operating at 50Hz.

For baseline comparisons, we implement state-of-the-art methods, Legolas, which was deployed on quadrupedal robots~\cite{17299a5987c541c3861795a72e1679bc} and demonstrated superiority over most other baseline methods such as filtering-based approaches. We also conduct extensive ablation studies on our method to validate that our approach achieves the optimal configuration and input modality combination.

We focus our comparative analysis on Legolas as the primary baseline because it has been extensively validated against traditional filtering methods (e.g., EKF-based approaches~\cite{DBLP:journals/firai/CamurriRNF20,10342061}) and hybrid learning-filtering methods (e.g., TLIO~\cite{9134860}) in prior work~\cite{17299a5987c541c3861795a72e1679bc}, demonstrating superior performance across multiple metrics. Including these additional baselines would be redundant given Legolas's established dominance in the literature. Therefore, outperforming Legolas provides sufficient evidence of our method's effectiveness relative to the broader landscape of proprioceptive odometry approaches.

Our evaluation follows standard odometry metrics~\cite{6096039}. We report Absolute Trajectory Error (ATE) in two variants: $ATE_o$ (first-frame alignment) and $ATE_u$ (Umeyama alignment), and Relative Pose Error (RPE) for local accuracy measurement.

Our evaluation follows standard odometry metrics~\cite{6096039}. We report Absolute Trajectory Error (ATE) in two variants: $ATE_o$ (first-frame alignment) and $ATE_u$ (Umeyama alignment), and Relative Pose Error (RPE) for local accuracy measurement.

RPE measures local incremental errors without cumulative drift:
\begin{equation}
RPE_{rmse} := \sqrt{\frac{1}{n} \sum_{i=1}^{n} \|\Delta P_{real} - \Delta P_{pred}\|^2}
\end{equation}

ATE measures global trajectory alignment error:
\begin{equation}
ATE_{rmse} := \sqrt{\frac{1}{n} \sum_{i=1}^{n} \|P_{real} - S P_{pred}\|^2}
\end{equation}

where $S$ is the alignment transformation matrix. $ATE_o$ uses first-frame alignment while $ATE_u$ uses Umeyama optimal alignment.

\subsection{Quantitative Analysis and Ablation Study}

\begin{table}[t]
\centering
\renewcommand{\arraystretch}{1.2} 
\setlength{\tabcolsep}{6pt} 
\caption{Performance comparison of different odometry methods. Best results are highlighted in bold.}
\label{tab:main_results}
\begin{tabular}{|c|c|c|c|c|}
\hline
\# & \textbf{Method} & \(\mathbf{ATE_o}\) & \(\mathbf{ATE_u}\) & \textbf{RPE} \\
\hline
1 & AutoOdom, Only Stage1 & 0.8176 & 0.2165 & 0.0124 \\
2 & AutoOdom, Only Stage1, LSTM & 0.9457 & 0.3666 & 0.0221 \\
4 & AutoOdom, Only Stage2 & 0.8016 & 0.3328 & 0.0185 \\
5 & AutoOdom, Only Stage2, LSTM & 0.6472 & 0.3404 & 0.0153 \\
6 & AutoOdom & \textcolor{red}{0.5630} & \textcolor{red}{0.2153} & \textcolor{red}{0.0148} \\
7 & AutoOdom, LSTM & \textcolor{red}{0.6196} & \textcolor{red}{0.2544} & \textcolor{red}{0.0153} \\
8 & Legolas & 1.3153 & 0.5281 & 0.0232 \\
9 & AutoOdom, w/o accel & 0.6861 & 0.2678 & 0.0162 \\
10 & AutoOdom, w/o autoregressive & 0.7876 & 0.2095 & 0.0141 \\
\hline
\end{tabular}
\end{table}

Table~\ref{tab:main_results} presents a comprehensive performance comparison between our AutoOdom method and the state-of-the-art Legolas baseline on our test dataset. The results demonstrate that AutoOdom achieves substantially superior performance across all evaluation metrics.

The experimental results reveal several critical insights into the effectiveness of our two-stage training framework. First, our complete AutoOdom system (row 6) achieves the best overall performance with $ATE_o$ of 0.5630, $ATE_u$ of 0.2153, and RPE of 0.0148, representing significant improvements of 57.2\% in $ATE_o$, 59.2\% in $ATE_u$, and 36.2\% in RPE compared to the Legolas baseline (row 8). Second, the comparison between single-stage and two-stage approaches validates our design philosophy: while Stage 1 alone (rows 1-2) provides reasonable performance through simulation-based pre-training, and Stage 2 alone (rows 4-5) shows the importance of real-world data, the complete two-stage framework (rows 6-7) achieves optimal results by effectively combining both advantages. Third, the architectural comparison between MLP and LSTM variants reveals that both architectures achieve nearly identical performance across all metrics, with only marginal differences in accuracy. Given this comparable performance and considering the computational efficiency requirements for real-time deployment, we ultimately adopt the MLP architecture due to its lower inference latency and reduced memory footprint compared to LSTM networks. The comprehensive comparison with the Legolas baseline validates the effectiveness of our autoregressive enhancement approach. AutoOdom consistently outperforms Legolas across all evaluation metrics, achieving substantial reductions in trajectory error while maintaining computational efficiency suitable for real-time deployment. 

To understand the contribution of different input modalities and architectural choices, we conduct comprehensive ablation studies examining the impact of various sensor inputs and temporal configurations. These ablation experiments are performed using only Stage 1 (simulation-based pre-training) to isolate the effects of individual components before the autoregressive enhancement phase. Table~\ref{tab:ablation} provides valuable insights into the system design choices and their individual contributions to overall performance.

\begin{table}[t]
\centering
\renewcommand{\arraystretch}{1.2} 
\setlength{\tabcolsep}{6pt} 
\caption{Ablation study on input modalities and configurations in stage 1}
\label{tab:ablation}
\begin{tabular}{|c|c|c|c|c|c|c|c|}
\hline
\# & \textbf{\(\Delta t\)} & \textbf{\(a_t\)} & \textbf{\(p_t\)} & \textbf{\(A_t\)} & \(\mathbf{ATE_o}\) & \(\mathbf{ATE_u}\) & \textbf{RPE} \\ 
\hline
1 & 0.02 & \checkmark & \checkmark & \checkmark & 1.4486 & 0.6577 & 0.0378 \\
2 & 0.02 & × & \checkmark & \checkmark & 0.8176 & 0.2165 & 0.0124 \\
3 & 0.02 & \checkmark & × & \checkmark & 1.1113 & 0.4840 & 0.0280 \\
4 & 0.02 & × & × & \checkmark & 0.8876 & 0.2195 & 0.0141 \\
5 & 0.02 & \checkmark & \checkmark & × & 1.9190 & 0.6739 & 0.0439 \\
6 & 1.02 & \checkmark & \checkmark & \checkmark & 1.0831 & 0.4689 & 0.1551 \\
\hline
\end{tabular}
\end{table}

\begin{figure*}[t]
    \centering
    \begin{minipage}{0.48\textwidth}
        \centering
        \includegraphics[width=\textwidth]{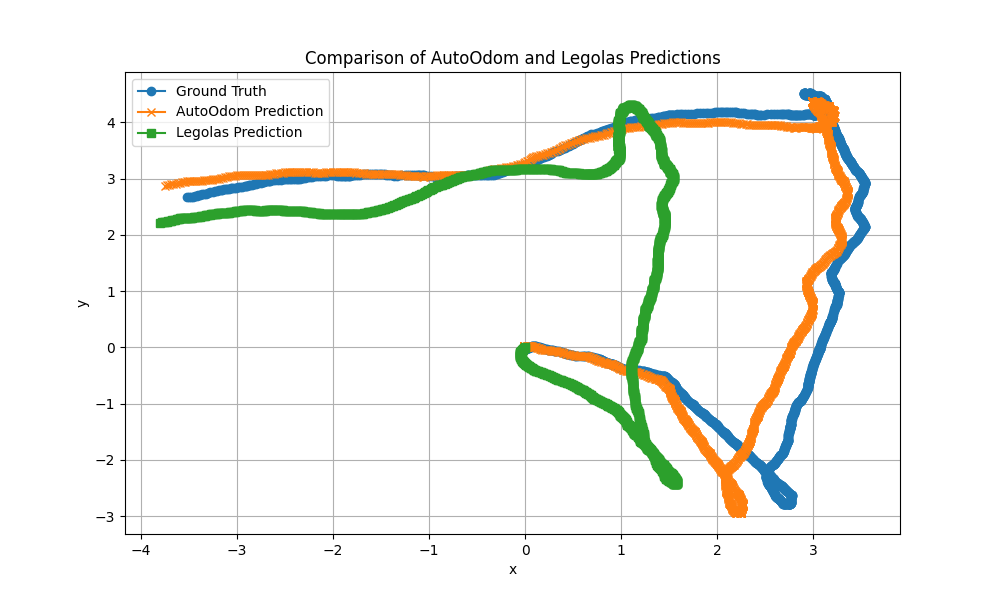}
        \caption*{(a) Complex trajectory with sharp turns}
    \end{minipage}
    \hfill
    \begin{minipage}{0.48\textwidth}
        \centering
        \includegraphics[width=\textwidth]{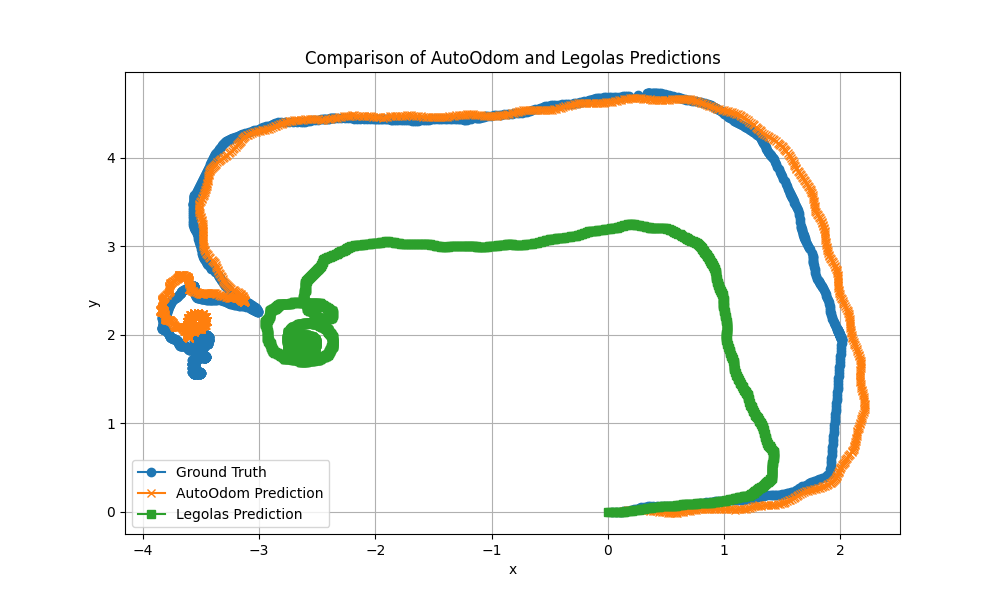}
        \caption*{(b) Rectangular path with precise corners}
    \end{minipage}
    \caption{Trajectory comparison between AutoOdom and Legolas on some test sequences. Blue: ground truth, orange: AutoOdom, green: Legolas. AutoOdom shows better alignment with ground truth, especially during directional changes.}
    \label{fig:trajectory_comparison}
\end{figure*}

\section{Conclusion}

The ablation study results in Table~\ref{tab:ablation} reveal key insights into input modality contributions. Notably, excluding IMU acceleration during Stage 1 training improves $ATE_o$ from 1.4486 to 0.8176, while incorporating it in Stage 2 enhances performance, highlighting the importance of stage-specific sensor configuration. Action history proves critical, with its removal causing $ATE_o$ to deteriorate to 1.9190, while historical position data and high-frequency prediction intervals (0.02s vs 1.02s) also contribute significantly to accuracy.

In this work, we present AutoOdom, a novel autoregressive proprioceptive odometry system that bridges the sim-to-real gap through a two-stage training approach. Our key contributions include: (1) a simulation pre-training followed by real-world fine-tuning scheme, and (2) an autoregressive mechanism enabling the model to learn from its own predictions. Experimental results demonstrate 57.2\% improvement in $ATE_o$, 59.2\% in $ATE_u$, and 36.2\% in RPE compared to state-of-the-art methods, with successful deployment on the Booster T1 humanoid robot.

Future work will extend the system to diverse environmental conditions, integrate uncertainty estimation for SLAM applications, and explore deployment on additional robotic platforms.

\section*{Acknowledgment}

This work was supported by the Department of Automation, Tsinghua University. We thank Booster Robotics for providing the Booster T1 humanoid robot platform and technical support. We also thank the anonymous reviewers for their valuable feedback.

\bibliographystyle{ieeetr}

\bibliography{reference.bib}

\end{document}